\documentclass{article}
\usepackage{spconf,amsmath,graphicx}

\usepackage{enumitem}
\setlist{nosep, leftmargin=14pt}

\usepackage{mwe} 
\usepackage[ruled,vlined]{algorithm2e}
\usepackage{algpseudocode}
\usepackage{booktabs} 
\usepackage{xcolor,colortbl}
\usepackage{array}        
\usepackage{makecell}     
\usepackage{xcolor}      
\usepackage{caption}     
\usepackage{graphicx}     
\usepackage{multirow}    
\usepackage{amssymb}
\usepackage{pifont}
\usepackage{xcolor,colortbl}
\usepackage[hidelinks]{hyperref}
\newcommand{\grayref}[1]{\textcolor{black!50}{\small #1}}
\DeclareMathOperator*{\argmax}{argmax}
\title{SAM-Fed: SAM-Guided Federated Semi-Supervised Learning for Medical Image Segmentation}

\name{
\begin{tabular}{c}
Sahar Nasirihaghighi$^{\star}$ 
Negin Ghamsarian$^{\dagger}$ 
Yiping Li$^{\ddagger}$
Marcel Breeuwer$^{\ddagger}$ \\
Raphael Sznitman$^{\dagger}$
Klaus Schoeffmann$^{\star}$ 
\end{tabular}
}

\address{
$^{\star}$ Institute of Information Technology (ITEC), University of Klagenfurt, Austria \\
$^{\dagger}$ Center for Artificial Intelligence in Medicine, University of Bern, Switzerland \\
$^{\ddagger}$ Department of Biomedical Engineering, Eindhoven University of Technology, Netherlands
}

\begin{document}

\maketitle
\begin{abstract}
Medical image segmentation is clinically important, yet data privacy and the cost of expert annotation limit the availability of labeled data. Federated semi-supervised learning (FSSL) offers a solution but faces two challenges: pseudo-label reliability depends on the strength of local models, and client devices often require compact or heterogeneous architectures due to limited computational resources. These constraints reduce the quality and stability of pseudo-labels, while large models, though more accurate, cannot be trained or used for routine inference on client devices. We propose SAM-Fed, a federated semi-supervised framework that leverages a high-capacity segmentation foundation model to guide lightweight clients during training. SAM-Fed combines dual knowledge distillation with an adaptive agreement mechanism to refine pixel-level supervision. Experiments on skin lesion and polyp segmentation across homogeneous and heterogeneous settings show that SAM-Fed consistently outperforms state-of-the-art FSSL methods.

\end{abstract}
\vspace{-0.5em}
\section{Introduction}
\label{sec:intro}
\vspace{-0.5em}
Deep learning has achieved strong performance in medical image analysis, supported by large labeled datasets~\cite{zhang2025federated}. However, sharing medical data across institutions is limited by privacy regulations and ethical concerns~\cite{mcmahan2017communication}. Federated Learning (FL) enables collaborative model training without exchanging raw data: each client trains locally and only shares model updates, which are then aggregated on a central server (e.g., using FedAvg~\cite{mcmahan2017communication}).

Most FL methods assume fully labeled local datasets, which is rarely feasible in clinical settings due to the cost of expert annotation~\cite{wu2023federated}. Since unlabeled medical images are far more abundant, Federated Semi-Supervised Learning (FSSL) has emerged as a more practical alternative~\cite{liang2022rscfed}. Existing FSSL approaches improve collaboration by promoting inter-client consistency~\cite{jeong2021federated}, leveraging global and local knowledge distillation~\cite{kim2024federated}, or employing data-free distillation and dynamic aggregation to handle heterogeneous client distributions~\cite{zhang2022fine, shen2025dynamic}. Recently, Ma et al.~\cite{mainbaseline} introduced HSSF, a model-heterogeneous FSSL framework that supports client-specific architectures and transfers knowledge through Regularity Condensation and Fusion.

Despite recent progress in FSSL, ensuring the reliability of pseudo-labels in semi-supervised semantic segmentation remains a central challenge~\cite{nasirihaghighi2025dual,ghamsarian2023domain}, as their quality depends strongly on the accuracy and capacity of each client’s local model. In principle, higher-capacity models can generate more precise segmentation predictions and thus more reliable pseudo-labels. However, in federated medical environments, client devices typically operate under strict computational constraints, making it impractical for them to train or sustain such large models for local learning and inference~\cite{mora2024knowledge}. This mismatch between the need for strong pseudo-label supervision and the limited capability of client-side models further complicates semi-supervised learning in federated settings.
To address this, we propose a dual knowledge distillation strategy composed of (I) federated knowledge distillation, which promotes bidirectional knowledge transfer between the global server model and local client models, and (II) SAM-guided knowledge distillation, which uses the Segment Anything Model (SAM) to provide fine-grained pixel-level supervision for lightweight local models. To the best of our knowledge, no prior work performs joint server–client knowledge distillation for segmentation in a fully unlabeled client scenario. Our main contributions are summarized as follows:

\begin{itemize}
    \item We propose a novel framework that leverages high computational resources available at the server side to enhance pseudo-supervision in unlabeled client-side settings.
    \item We propose a semi-supervised semantic segmentation strategy based on pixel-level prediction agreement between the server-side trained teacher and the client-side model, enabling reliable pseudo-label generation.
    \item Our framework supports both homogeneous aggregation via FedAvg and heterogeneous aggregation, allowing flexible deployment across clinical environments with varying computational resources.
\end{itemize}

\begin{figure}[!t]
    \centering
    \captionsetup{font=small}
    \includegraphics[width=0.48\textwidth]{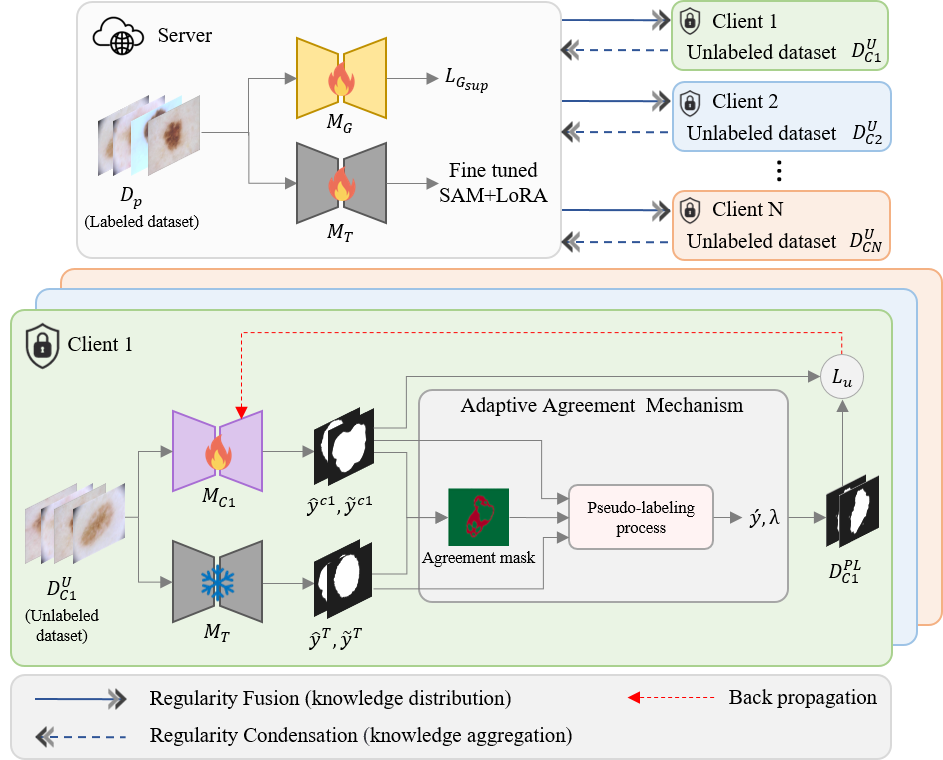}
    \caption{Overview of SAM-Fed.}
    \label{fig:Overview}
\end{figure}

\section{Proposed Approach}
\label{sec:approach}
\vspace{-0.5em}
\noindent\textbf{Problem Definition.}
Consider a federated learning setup consisting of a central server and $c \in \{C1, C2, \ldots, CN\}$ clients. The server hosts a public labeled dataset $\mathcal{D}_{p}={\{(x^p_i, y^p_i)\}_{i=1}^{N_p}}$ containing $\mathcal{N}_{p}$ pairs, while each client $c$ owns a private unlabeled dataset $\mathcal{D}_{c}={\{x^c_j\}_{j=1}^{N_c}}$, with $\mathcal{N}_{c}$ images. The data distributions across clients are non-identical and heterogeneous. 
The goal is to collaboratively train local segmentation models through a semi-supervised process that utilizes publicly shared labeled data, alongside private unlabeled data, on each client.

\textbf{Overview.}  
The overall architecture of the proposed framework in a heterogeneous client setting is illustrated in Fig.~\ref{fig:Overview}. Our proposed framework consists of three main components: (1) server-side supervised training, (2) client-side SSL with adaptive agreement-based pseudo-labeling, and (3) federated aggregation. Our approach is model-agnostic and can be integrated into various federated semi-supervised frameworks. Here, we use HSSF model as a baseline and tackle the issue of unreliable pseudo-labels, which often results in unstable supervision and diminished segmentation performance. Further implementation details of the baseline framework can be found in~\cite{mainbaseline}.

\begin{table*}[!ht]
\renewcommand{\arraystretch}{0.9}
\centering
\small
\caption{Comparison with state-of-the-art FSSL methods for skin lesion segmentation, evaluated on local datasets.}\vspace{-0.5em}
\label{tab:SAMFed_isic}

\resizebox{\textwidth}{!}{
\begin{tabular}{
    >{\raggedright\arraybackslash}m{2.2cm}   
    >{\centering\arraybackslash}m{2.3cm}   
    >{\centering\arraybackslash}m{1.0cm}   
    *{5}{>{\centering\arraybackslash}m{1.2cm}} 
    *{5}{>{\centering\arraybackslash}m{1.2cm}} 
}
\toprule
\multirow{2}{*}{Framework} &
\multirow{2}{*}{} &
\multirow{2}{*}{Hetero.} &
\multicolumn{5}{c}{Dice (\%) $\uparrow$} &
\multicolumn{5}{c}{HD95 (\%) $\downarrow$} \\
\cmidrule(lr){4-8}\cmidrule(lr){9-13}
 & & & C1 & C2 & C3 & C4 & Avg. & C1 & C2 & C3 & C4 & Avg. \\
\midrule
MeanTeacher~\cite{meanteacher}& \grayref{[NeurIPS 2017]} & \ding{55} & 69.39 & 69.76 & 73.59 & 77.29 & 74.39 & 66.39 & 54.49 & 48.92 & 48.01 & 50.70 \\
UniMatch~\cite{UniMatch}& \grayref{[CVPR 2023]} & \ding{55} & 76.81 & 78.54 & 80.26 & 81.11 & 80.12 & 51.30 & 41.03 & 34.80 & 37.15 & 38.13 \\
ELN~\cite{ELN}& \grayref{[CVPR 2022]} & \ding{55} & 73.05 & 76.16 & 72.10 & 75.90 & 74.55 & 38.35 & 39.85 & 48.21 & 43.24 & 43.88 \\
SemiFL~\cite{semiFL}& \grayref{[NeurIPS 2022]} & \ding{55} & 77.13 & 78.09 & 79.52 & 82.71 & 80.59 & 50.48 & 46.52 & 45.20 & 37.87 & 42.42 \\
LSSL~\cite{mainbaseline}& \grayref{[TMI 2024]} & \ding{55} & 86.59 & 81.06 & 81.47 & 84.94 & 83.40 & 19.35 & 38.28 & 29.81 & 27.43 & 29.21 \\
\arrayrulecolor{gray!60}\midrule\arrayrulecolor{black}
SAM-Fed (Ours) & &  \ding{55} & \textbf{87.41} & \textbf{86.54} & \textbf{86.28} & \textbf{87.52} & \textbf{86.98} & \textbf{18.96} & \textbf{29.58} & \textbf{25.52} & \textbf{23.41} & \textbf{24.67} \\


\arrayrulecolor{gray!60}\midrule\arrayrulecolor{black}

FedMD~\cite{li2019fedmd}& \grayref{[NeurIPS 2019]} & \ding{51} & 70.41 & 70.94 & 73.17 & 77.86 & 74.78 & 49.43 & 47.22 & 52.20 & 42.75 & 46.84 \\
KT-pFL~\cite{KT-pFL}& \grayref{[NeurIPS 2021]} & \ding{51} & 58.18 & 51.52 & 50.91 & 65.35 & 58.23 & 68.12 & 80.81 & 138.6 & 62.86 & 89.24 \\
FedProto~\cite{FedProto}& \grayref{[AAAI 2022]} & \ding{51} & 68.01 & 76.95 & 69.40 & 75.92 & 73.46 & 56.04 & 50.14 & 58.60 & 60.95 & 58.19 \\
RHFL~\cite{RHFL}& \grayref{[CVPR 2022]} & \ding{51} & 65.89 & 67.92 & 73.43 & 77.12 & 73.70 & 44.57 & 48.69 & 46.92 & 38.53 & 43.12 \\
HSSF~\cite{mainbaseline}& \grayref{[TMI 2024]} & \ding{51} & 72.18 & \textbf{77.91} & 77.34 & 81.83 & 79.11 & 51.10 & 41.74 & 32.30 & \textbf{30.35} & 34.18 \\
\arrayrulecolor{gray!60}\midrule\arrayrulecolor{black}

SAM-Fed (Ours) & & \ding{51} & \textbf{78.28} & 76.52 & \textbf{85.77} & \textbf{85.54} & \textbf{83.66} & \textbf{40.03} & \textbf{37.91} & \textbf{28.29} & 31.55 & \textbf{32.16} \\


\bottomrule
\end{tabular}
}
\end{table*}

\textbf{Server Side.}
At the initialization phase, the server trains a supervised global model on the public labeled dataset $\mathcal{D}_{p}$ and a high-capacity teacher model on the same dataset. While any high-performance model could serve as the teacher in our framework, we adopt SAM for its strong generalization capability, denoted as $\mathcal{M}_{T}$. We fine-tune SAM on the server side using the labeled dataset and Low-Rank Adaptation (LoRA), and then distribute the resulting teacher model, along with the public labeled dataset $\mathcal{D}_{p}$, to all clients for pseudo-supervision. On each client, this teacher distills knowledge into a lightweight semi-supervised model through an adaptive, agreement-driven mechanism.

\begin{algorithm}[t]
\footnotesize
\DontPrintSemicolon
\SetKwInOut{Input}{Input}
\SetKwInOut{Output}{Output}
\caption{SAM-Fed Training}
\label{alg:hssf_sam}

\textbf{Server side}\;
\Input{Public dataset $\mathcal{D}_{p}$, SAM teacher $\mathcal{M}_{T}$, global model $\mathcal{M}_{G}$}

Fine-tune $\mathcal{M}_{T}$ on $\mathcal{D}_{p}$, Train $\mathcal{M}_{G}$ on $\mathcal{D}_{p}$\;
Broadcast the weights of $\mathcal{M}_{T}$ and $\mathcal{D}_{p}$ to all clients\;

\vspace{0.5ex}
\textbf{Client side}\;
\Input{Public dataset $\mathcal{D}_{p}$, local unlabeled dataset $\mathcal{D}^{U}_{c}$, client model $\mathcal{M}_{c}$, $\mathcal{D}^{PL}_{c} = \varnothing$}

Train local model $\mathcal{M}_{c}$ on $\mathcal{D}_{p}$\;

\For{$\mathbf{x}_{j}^{c} \in \mathcal{D}^{U}_{c}$}{

    Generate pseudo-labels using $\mathcal{M}_{T}$ and $\mathcal{M}_{c}$\;

    Generate pixel-wise agreement mask using the Adaptive Agreement Mechanism 
    \Comment{Eqs.~\eqref{eq:predictions}--\eqref{eq:pseudo_label}}\;

    Compute confidence weights $\lambda_{j}$ for each pixel \Comment{Eqs.~\eqref{eq:confidence_weight}}\;

    $\mathcal{D}^{PL}_{c} \leftarrow \acute{y}_j, \lambda_{j}$\;
}

Train $\mathcal{M}_{c}$ on $\mathcal{D}^{PL}_{c}$ using confidence-weighted consistency loss. \;
Apply federated aggregation.

\end{algorithm}

\textbf{Client Side.}
Each client $c$ performs local training using SSL with adaptive agreement-based pseudo-label generation.
The local model $\mathcal{M}_c$ is initially trained on the public labeled dataset $\mathcal{D}_p$. This ensures the model establishes a strong foundation by learning from labeled data, which prepares it before attempting to leverage the client's private, unlabeled data. It is important to note that SAM is used to generate initial pseudo-labels only once through a single inference pass, which is computationally lightweight compared to training, and these predictions are then dynamically ensembled with the client model’s outputs during semi-supervised learning.

\textbf{Adaptive Agreement Mechanism:}
During early training, the lightweight client model typically produces low-confidence predictions on unlabeled data, whereas the teacher provides more reliable guidance. As training progresses, the client model becomes better aligned with its local data distribution, and its predictions gradually become more reliable than those of the frozen teacher. To account for this shift, we introduce an adaptive pseudo-labeling strategy that dynamically adjusts the contribution of teacher and client predictions at each iteration based on their relative confidence levels.

At each training iteration, both the client model and the frozen teacher predict labels for each sample $\mathbf{x}_{j}^c$ in the unlabeled batch:
\begin{equation}
\begin{cases}
\hat{y}^T_j = \argmax_{n\in N}\big(\tilde{y}^T_j\big), & \tilde{y}^T_j = \sigma\big(\mathcal{M}_T(x^c_j)\big)
\\
\hat{y}^c_j = \argmax_{n\in N}\big(\tilde{y}^c_j\big), & \tilde{y}^c_j = \sigma\big(\mathcal{M}_c(x^c_j)\big)
\end{cases}
\label{eq:predictions}
\end{equation}
where $\tilde{y}^T_j$ and $\tilde{y}^c_j$ denote the softmax outputs of the teacher and client models, respectively, and $\hat{y}^T_j$ and $\hat{y}^c_j$ are the corresponding predicted class labels.

The final pseudo-label for each pixel location $(w,h)$ is determined by an agreement-aware selection strategy. When the teacher and client predictions agree, we accept their consensus. Otherwise, we select the prediction with higher confidence:
\begin{equation}
\acute{y}_j(w,h) =
\begin{cases}
\hat{y}^T_j(w,h) = \hat{y}^c_j(w,h), & \text{if } \hat{y}^T_j(w,h) = \hat{y}^c_j(w,h)
\\
\hat{y}^T_j(w,h), & \text{if } s^T > s^c
\\
\hat{y}^c_j(w,h), & \text{otherwise}
\end{cases}
\label{eq:pseudo_label}
\end{equation}
where the first case explicitly captures model agreement, reinforcing reliable predictions through consensus.

To weight the confidence of each pseudo-label, we define an adaptive confidence weight $\lambda(w,h)$. Let $s^T = \max\big(\tilde{y}^T_j(w,h)\big)$ and $s^c = \max\big(\tilde{y}^c_j(w,h)\big)$ denote the maximum confidence scores of the teacher and client, respectively. Then:
\begin{equation}
\lambda_j(w,h) =
\begin{cases}
1, & \hat{y}^T_j(w,h) = \hat{y}^c_j(w,h)
\\
s^T, & s^T > s^c
\\
s^c, & \text{otherwise}
\end{cases}
\label{eq:confidence_weight}
\end{equation}
This weighting scheme assigns full confidence when both models agree, and otherwise uses the maximum confidence score from the more confident model.

After generating the pseudo-labeled dataset $\mathcal{D}_c^{PL} = \{(\acute{y}_j,\lambda_j)\}_{j=1}^{N_c}$ with adaptive agreement masks, the local model is further trained on the unlabeled data using the generated pseudo-labels and their associated weights. The unsupervised loss for the unlabeled data is computed as a weighted cross-entropy loss:
\begin{equation}
\mathcal{L}_u = \frac{1}{|B_u|} \sum_{j \in B_u} \sum_{w,h} \lambda_j(w,h) \cdot \ell_{CE}\big( \hat{y}_j(w,h), \acute{y}_j(w,h)\big)
\label{eq:unsupervised_loss}
\end{equation}
where $B_u$ denotes the batch of unlabeled samples, $\ell_{CE}$ is the cross-entropy loss, and $\lambda(w,h)$ modulates the contribution of each pixel based on the prediction confidence. This phase enables the model to exploit abundant unlabeled data while prioritizing reliable pseudo-label regions. Through iterative federated updates, both local and global models improve, progressively refining pseudo-label quality over communication rounds. Algorithm \ref{alg:hssf_sam} provides a detailed step-by-step overview of this process.

\begin{table*}[!ht]
\renewcommand{\arraystretch}{0.9}
\centering
\small
\caption{Comparison with state-of-the-art FSSL methods for polyp lesion segmentation, evaluated on local datasets.}\vspace{-0.5em}
\label{tab:SAMFed_polyp}

\resizebox{\textwidth}{!}{
\begin{tabular}{
    >{\raggedright\arraybackslash}m{2.2cm}   
    >{\centering\arraybackslash}m{2.3cm}   
    >{\centering\arraybackslash}m{1.0cm}   
    *{5}{>{\centering\arraybackslash}m{1.2cm}} 
    *{5}{>{\centering\arraybackslash}m{1.2cm}} 
}
\toprule
\multirow{2}{*}{Framework} &
\multirow{2}{*}{} &
\multirow{2}{*}{Hetero.} &
\multicolumn{5}{c}{Dice (\%) $\uparrow$} &
\multicolumn{5}{c}{HD95 (\%) $\downarrow$}\\
\cmidrule(lr){4-8}\cmidrule(lr){9-13}
 & & & C1 & C2 & C3 & C4 & Avg. & C1 & C2 & C3 & C4 & Avg. \\
\midrule
MeanTeacher~\cite{meanteacher}& \grayref{[NeurIPS 2017]} & \ding{55} & 17.23 & 17.44 & 11.04 & 27.88 & 21.06 & 250.4 & 249.7 & 275.5 & 207.1 & 234.9 \\
UniMatch~\cite{UniMatch}& \grayref{[CVPR 2023]} & \ding{55} & 35.37 & 34.34 & 71.27 & 52.32 & 47.14 & 83.77 & 113.2 & 26.71 & 76.22 & 76.21 \\
ELN~\cite{ELN}& \grayref{[CVPR 2022]} & \ding{55} & 40.37 & 27.45 & 37.56 & 53.10 & 43.27 & 92.66 & 106.5 & 66.95 & 90.14 & 90.70 \\
SemiFL~\cite{semiFL}& \grayref{[NeurIPS 2022]} & \ding{55} & 54.59 & 42.56 & 73.86 & 65.21 & 59.59 & 97.81 & 119.6 & 65.17 & 84.00 & 91.30 \\
LSSL~\cite{mainbaseline}& \grayref{[TMI 2024]} & \ding{55} & 56.79 & 56.07 & 74.16 & 65.88 & 62.83 & 66.00 & 60.73 & \textbf{27.36} & 63.94 & 59.20 \\
\arrayrulecolor{gray!60}\midrule\arrayrulecolor{black}
SAM-Fed (Ours) & & \ding{55} & \textbf{57.56} & \textbf{73.69} & \textbf{77.40} & \textbf{72.16} & \textbf{69.20} & \textbf{56.83} & \textbf{37.65} & 33.26 & \textbf{41.85} & \textbf{52.81} \\

\arrayrulecolor{gray!60}\midrule\arrayrulecolor{black}

FedMD~\cite{li2019fedmd}& \grayref{[NeurIPS 2019]} & \ding{51} & 52.47 & 39.81 & 71.58 & 36.34 & 45.81 & 90.38 & 70.61 & 37.13 & 110.6 & 89.68 \\
KT-pFL~\cite{KT-pFL}& \grayref{[NeurIPS 2021]} & \ding{51} & 58.37 & 32.22 & 72.38 & 52.60 & 53.19 & 70.88 & 117.3 & 34.65 & 84.12 & 79.11 \\
FedProto~\cite{FedProto}& \grayref{[AAAI 2022]} & \ding{51} & 39.44 & 22.47 & 47.36 & 51.19 & 42.60 & 108.6 & 110.5 & 41.85 & 127.2 & 109.8 \\
RHFL~\cite{RHFL}& \grayref{[CVPR 2022]} & \ding{51} & 56.83 & 40.03 & 68.71 & 63.12 & 58.18 & 68.36 & 86.26 & 37.77 & 69.47 & 68.00\\
HSSF~\cite{mainbaseline}& \grayref{[TMI 2024]} & \ding{51} & \textbf{66.35} & 42.73 & 78.62 & \textbf{72.82} & 66.65 & 51.07 & 83.24 & \textbf{25.30} & 54.15 & 54.01 \\
\arrayrulecolor{gray!60}\midrule\arrayrulecolor{black}
SAM-Fed (Ours) & & \ding{51} & 64.85 & \textbf{69.65} & \textbf{78.94} & 72.17 & \textbf{68.97} & \textbf{47.05} & \textbf{59.73} & 44.62 & \textbf{53.04} & \textbf{51.11} \\
\bottomrule
\end{tabular}
}
\end{table*}


\textbf{Federated Aggregation.}
We adopt different aggregation strategies for homogeneous and heterogeneous client settings. In the homogeneous case, where all clients share the same architecture, we use FedAvg to compute a data-weighted average of local parameters after each round and broadcast the updated global model. For the heterogeneous case, we follow the HSSF baseline~\cite{mainbaseline} using Regularity Condensation (RC) and Regularity Fusion (RF): the server extracts reliable knowledge from client predictions on the public dataset $\mathcal{D}_{p}$ (RC), then returns this refined knowledge to clients, which incorporate it via a KL-divergence loss (RF), enabling collaboration across different architectures.

\section{Experimental Settings}
\label{sec:settings}
\vspace{-0.5em}
\textbf{Dataset.}
We evaluate our framework on two tasks: skin lesion and polyp segmentation. For skin lesions, we use ISIC2018 (2,694 images), reserving 100 images as public data and distributing the remaining samples across four clients with 200, 400, 800, and 1,194 images ($\mathcal{D}{1}$–$\mathcal{D}{4}$). For polyps, we adopt a Non-IID setting using five datasets: CVC-ColonDB (public, 380 images) and four private client datasets, CVC-ClinicDB (300 images), EndoTect-ETIS (612), CVC-300 (396), and Kvasir (1,000). Each private dataset is split into training, validation, and test sets with a 0.8:0.1:0.1 ratio.

\noindent \textbf{Networks and Training Settings.}
We simulate four client sites under both homogeneous and heterogeneous settings. In the heterogeneous configuration, $\mathcal{C}{1}$ and $\mathcal{C}{2}$ use ResNet18 backbones, while $\mathcal{C}{3}$ and $\mathcal{C}{4}$ use ResNet34; the server employs ResNet101 for feature extraction. In the homogeneous setting, all clients use ResNet34. All images are resized to 384×384 and processed using the same augmentation and hyperparameter settings as~\cite{mainbaseline}. Training uses AdamW optimizer with an initial learning rate of 0.0001. For SAM fine-tuning on the server side, both the vision and prompt encoders are frozen for efficiency. SAM is adapted via LoRA with rank $r=16$, scaling factor $\alpha=32$, dropout 0.1, resulting in 6.65M trainable parameters.

\section{Experimental Results}
\label{sec:results}
\vspace{-0.5em}
Tables~\ref{tab:SAMFed_isic} and~\ref{tab:SAMFed_polyp} report the performance of SAM-Fed on the ISIC2018 and polyp datasets under two configurations. The upper parts correspond to the homogeneous setting, where all clients use identical architectures and aggregation follows FedAvg, enabling comparison with FSSL methods (SemiFL, LSSL) and semi-supervised baselines (MeanTeacher, UniMatch). The lower parts reflect the heterogeneous setting, where clients employ different backbone architectures but identical data partitions, and SAM-Fed is compared against heterogeneous FL frameworks such as FedMD, KT-pFL, FedProto, RHFL, and HSSF.
On ISIC2018, SAM-Fed achieves superior performance in both settings. In the homogeneous case, it attains the highest average Dice (86.98\%) and lowest HD95 (24.67\%), surpassing all baselines. Under heterogeneous conditions, it maintains the best overall Dice (83.66\%) and HD95 (32.16\%), despite minor variations across clients. For the polyp datasets, SAM-Fed similarly outperforms all baselines, achieving 69.20\% Dice and 52.81\% HD95 in the homogeneous setting and 68.97\% Dice and 51.11\% HD95 in the heterogeneous setting, demonstrating strong robustness and generalizability across diverse architectures and non-IID environments.

\begin{figure}[!t]
    \centering
    \captionsetup{font=small}
    \includegraphics[width=0.49\textwidth]{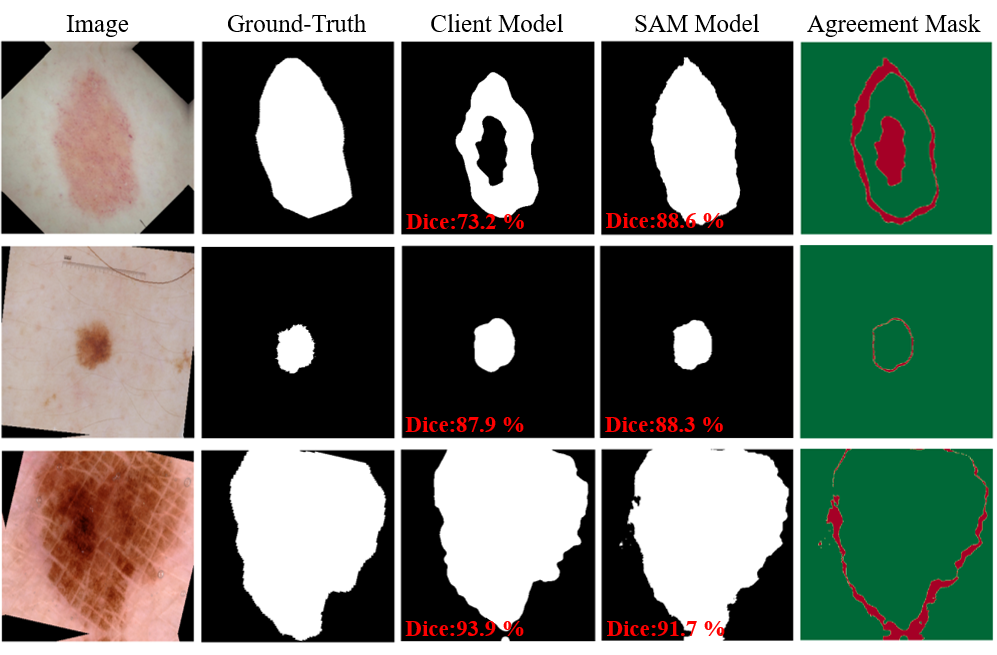}
    \caption{Adaptive agreement mechanism for pseudo-label generation, with examples from early (row 1), middle (row 2), and final (row 3) training epochs. The agreement mask highlights pixel-wise consistency between predictions (green = agreement, red = disagreement).}
    \label{fig:PseudoLabelGeneration}
\end{figure}

Fig.~\ref{fig:PseudoLabelGeneration} illustrates the pseudo-label generation process in the heterogeneous setting, showing the predictions of SAM and the client model together with their agreement masks, where green indicates agreement and red indicates disagreement. Each row corresponds to a different training epoch and highlights the progressive refinement of the client model. In the early stages, the client model benefits substantially from SAM’s stronger predictions, using them as reliable supervisory signals. As training progresses, the client model gradually becomes more accurate and increasingly aligned with SAM, eventually producing high-quality predictions even in challenging regions. This visual evolution underscores the role of the adaptive pixel-wise agreement mechanism in providing stable and informative supervision throughout training. Together with the quantitative results reported earlier, these observations confirm that SAM-Fed effectively leverages SAM’s generalization ability while enabling clients with heterogeneous architectures to converge toward robust and consistent segmentation performance.

\section{Conclusion}
\label{sec:Conclusion}
\vspace{-0.5em}
We proposed SAM-Fed, a SAM-guided federated semi-supervised learning framework designed to address computational constraints at clinical sites, where training large models locally is often infeasible. SAM-Fed introduces a dual knowledge distillation strategy that integrates federated knowledge distillation for bidirectional knowledge exchange between global and client models with SAM-guided distillation, which leverages the Segment Anything Model to provide fine-grained pixel-level supervision for lightweight clients. An adaptive pixel-level agreement mechanism further refines pseudo-labels dynamically, ensuring reliable and consistent supervision throughout training. Experiments on skin lesion and polyp segmentation demonstrate that SAM-Fed consistently outperforms state-of-the-art FSSL baselines in both homogeneous and heterogeneous configurations, validating its effectiveness and practicality for real-world federated medical image segmentation.

\bibliographystyle{ieeetr}
\bibliography{refs}

\end{document}